\documentclass[10pt,twocolumn,letterpaper]{article}

\usepackage{iccv}
\usepackage{times}
\usepackage{epsfig}
\usepackage{graphicx}
\usepackage{amsmath}
\usepackage{amssymb}
\usepackage{enumitem} 

\usepackage{algorithm}
\usepackage{algorithmic}
\usepackage{caption}
\usepackage{subcaption}
\usepackage{array}
\usepackage{colortbl}
\usepackage{booktabs}
\usepackage{multirow}
\usepackage{multicol}
\usepackage{makecell}
\usepackage{xcolor}
\usepackage{xspace}
\usepackage{float}
\usepackage{color}
\usepackage{bbding}
\usepackage{pifont}
\usepackage{enumitem} 
\usepackage{diagbox}
\usepackage{makecell}
\usepackage[accsupp]{axessibility} 

\newcommand{\myparagraph}[1]{{\vspace{.5em} \noindent \bf #1}}

\usepackage[pagebackref,breaklinks,colorlinks]{hyperref}

\iccvfinalcopy 

\def\iccvPaperID{****} 

\ificcvfinal\fi

\begin{document}

\title{Learning Cross-Modal Affinity for Referring Video Object Segmentation Targeting Limited Samples}

\author{
Guanghui Li$^{1}$\thanks{Equal contribution. This work was done when G. Li visited to Feng Zheng Lab in Southern University of Science and Technology.}\ ,
Mingqi Gao$^{2,3}$\footnotemark[1] ,
Heng Liu$^1$\thanks{Corresponding author.}\ ,
Xiantong Zhen$^4$,
Feng Zheng$^2$\footnotemark[2]
\\
$^1$ Anhui University of Technology,
$^2$ Southern University of Science and Technology,\\
$^3$ University of Warwick,
$^4$ United Imaging
\\
{\tt\small guanghui.li1998@gmail.com,}
{\tt\small mingqi.gao@outlook.com,}
{\tt\small hengliusky@aliyun.com,}\\
{\tt\small zhenxt@gmail.com,}
{\tt\small f.zheng@ieee.org}
}


\maketitle
\ificcvfinal\thispagestyle{empty}\fi

\begin{abstract}
Referring video object segmentation (RVOS), as a supervised learning task, relies on sufficient annotated data for a given scene. However, in more realistic scenarios, only minimal annotations are available for a new scene, which poses significant challenges to existing RVOS methods.
With this in mind, we propose a simple yet effective model with a newly designed cross-modal affinity (CMA) module based on a Transformer architecture. 
The CMA module builds multimodal affinity with a few samples, thus quickly learning new semantic information, and enabling the model to adapt to different scenarios.
Since the proposed method targets limited samples for new scenes, we generalize the problem as - few-shot referring video object segmentation (FS-RVOS).
To foster research in this direction, we build up a new FS-RVOS benchmark based on currently available datasets. The benchmark covers a wide range and includes multiple situations, which can maximally simulate real-world scenarios.
Extensive experiments show that our model adapts well to different scenarios with only a few samples, reaching state-of-the-art performance on the benchmark. On Mini-Ref-YouTube-VOS, our model achieves an average performance of 53.1 $\mathcal{J}$ and 54.8 $\mathcal{F}$, which are 10\% better than the baselines. Furthermore, we show impressive results of 77.7 $\mathcal{J}$ and 74.8 $\mathcal{F}$ on Mini-Ref-SAIL-VOS, which are significantly better than the baselines. 
Code is publicly available at \url{https://github.com/hengliusky/Few_shot_RVOS}. 

\end{abstract}


\section{Introduction}
\label{sec:intro}

\begin{figure}[t]
  \centering
  \includegraphics[width=0.99\linewidth]{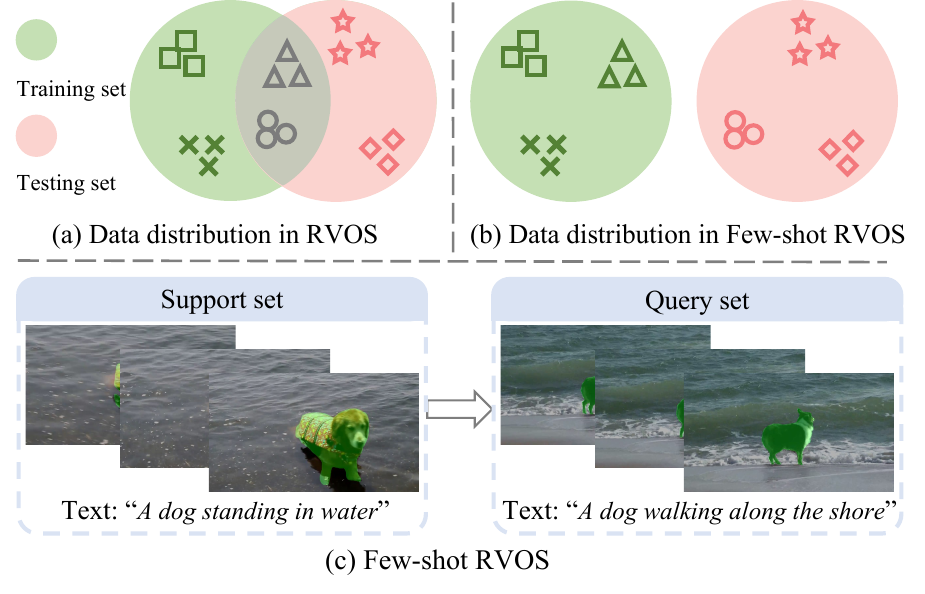}
   \caption{Comparison of Few-shot RVOS and RVOS and the setting of Few-shot RVOS.
(a) The training and testing sets overlap in the RVOS.
(b) Disjoint training and testing sets in the Few-shot RVOS. 
Different shapes represent different classes.
(c) Few-shot RVOS segments the referred object of the same class as the support set in the video.}
   \label{fig:figure1}
\end{figure}


Referring video object segmentation (RVOS) aims to segment target objects described in natural language in videos. In real-world scenarios, it has a wide range of applications, such as video editing~\cite{fu2022m3l} and human-computer interaction, so RVOS has attracted much attention from the research community. 
Unlike traditional semi-supervised video object segmentation~\cite{gao2023deep}, RVOS is more challenging because it not only lacks the ground-truth mask of the first frame of the video but also needs to interact with multimodal information of vision and language. \par

The great success of various tasks based on deep learning benefits from sufficient labeled data.
Detailed annotated masks and language descriptions in real-world RVOS tasks are relatively scarce. The researchers have to annotate each frame in the video in detail and provide a referring expression for the segmentation object. Therefore, obtaining high-quality labeled data requires a high cost.
With the popularity of movies, YouTube videos, TikTok streaming videos, etc., video data in various fields has shown explosive growth in this media age. 
The demand for processing diverse data has brought significant challenges.
In order to handle diverse data, existing RVOS methods must rely on massive and diverse labeled data for training.
But due to fixed and limited training classes, existing RVOS methods~\cite{botach2021mttr,wu2022language} are essentially constrained to adapt to the highly dynamic and highly diverse data in the real world.
On the other hand, if ones fine-tune existing RVOS methods on a few samples to adapt to real-world data, high-quality results are hard to be achieved because the labelled data is insufficient to support the model for learning the new semantics. 
Therefore, how to make the RVOS methods applicable to real-world diverse data with a lower cost is an urgent problem.
\par

To address this problem, we propose the cross-modal affinity (CMA) module to build multimodal relationships in a few samples and learn new semantic information for diversified data. Specifically, given only a few annotated samples (language expressions and the referred object masks), we hierarchically fuse visual and text features in a cross-attention manner to obtain robust feature representations for a specific category. In this way, the model can handle enormous data in the same category more efficiently. \par

Essentially, the proposed method targets limited samples. 
Therefore, we generalize the problem as Few-Shot Referring Video Object Segmentation (FS-RVOS).
We show the setting of FS-RVOS and the difference from existing RVOS in Figure~\ref{fig:figure1}. Unlike RVOS, the training and testing sets' categories disjoint in the FS-RVOS.
Given a few support video clips together with corresponding language descriptions and object masks, FS-RVOS aims at segmenting videos in the query set, as shown in Figure~\ref{fig:figure1}(c). \par

The key to FS-RVOS lies in the support set utilization and understanding of vision-language information. To better leverage the information in the support set, two methods have been proposed based on the prototype and attention mechanisms. 
The prototype-based~\cite{nguyen2019feature,yang2020prototype} methods compress the features belonging to different classes to obtain prototypes. However, noise is easily generated during the process. In addition, the spatial structures are ignored, resulting in different degree of information loss. Another methods~\cite{zhang2019pyramid,zhang2021few,zhang2022catrans} employ the attention mechanism to encode foreground pixels from support features and aggregate them with query features.
Although these methods achieve high-quality results in image and video domains, they are still under-explored in vision-language tasks. \par

To better utilize vision-language inputs, we propose the cross-modal affinity module to build the multimodal relationships between samples in the support and query sets. Specifically, multimodal features within the support set and query set are first fused separately. 
The information among them is then aggregated, which effectively prevents query features from being biased by irrelevant features.
\par

Since this is the first work exploring Few-shot RVOS, the existing datasets are not directly applicable. 
Therefore, we build up a new FS-RVOS benchmark based on Ref-YouTube-VOS~\cite{seo2020urvos}, named Mini-Ref-YouTube-VOS. 
The new benchmark covers a wide range with a balanced number of high-quality videos in each category.
To measure the model's generalization ability, we also build a dataset different from natural scenes based on a synthetic dataset SAIL-VOS~\cite{hu2019sail}, named Mini-Ref-SAIL-VOS. 
Since only videos and detailed annotated masks exist in the SAIL-VOS dataset, we add natural language descriptions corresponding to the segmentation targets for the dataset. \par


The main contributions of this work are as follows.
\begin{itemize}[leftmargin=*]
\vspace{-2mm}
\item For real-world limited samples, we propose a \textbf{C}ross-\textbf{M}odal \textbf{A}ffinity (CMA) for building multimodal information affinity for referring video object segmentation.
\vspace{-2mm}
\item  We explore a novel Few-shot RVOS problem, which learns new semantic information with limited samples and can adapt to diverse scenarios.
\vspace{-2mm}
\item We build up the first FS-RVOS benchmark, where we conduct comprehensive comparisons with existing methods, showing the superiority of the proposed model. 
\end{itemize} 

\section{Related Work}
\label{sec:related}
\myparagraph{Few-Shot Semantic Segmentation.} 
Few-shot semantic segmentation, first proposed by Shaban et al.~\cite{shaban2017one}, aims to learn how to segment new categories of images through a few samples. Recent advances in few-shot semantic segmentation originate from the application of metric learning. Based on PrototypicalNet, Dong et al.~\cite{dong2018few} first employ the metric learning technique and apply cosine similarity between pixels and prototypes for prediction.
In addition, PANet et al.~\cite{wang2019panet} introduce prototype alignment regularization to simplify the framework.
PFENet et al.~\cite{tian2020prior} use prior knowledge from the pretrained backbone to find the regions of interest and the different designs of feature pyramid modules, and previous leverage mappings to achieve better segmentation performance.\par


However, the effectiveness of these few-shot segmentation methods depends mainly on the quality of the prototypes obtained from the support set. Fan et al. \cite{fan2022ssp} address the critical intra-class appearance differences inherent in the few-shot segmentation problem by performing self-support matching with query features. Their strategy effectively captures the consistent underlying features of query objects to match query features. 
Tian et al. ~\cite{Tian_2022_CVPR} propose a novel context-aware prototype learning method that leverages prior knowledge from support samples and dynamically enriches contextual information by using adaptive features. Motivated by the idea that using the base learner to identify confusing regions in the query image and further refining the predictions of the meta-learner, BAM ~\cite{Lang_2022_CVPR} establishes a new method for few-shot segmentation that does not focus on feature extraction or visual correspondence.\par

Compared to image-based few-shot segmentation, few-shot video object segmentation works are relatively rare and remain in the early stage. The initial work ~\cite{siam2021weakly},~\cite{chen2021delving} mainly solves this problem through the attention mechanism.
However, these methods do not consider temporal information. Thus, based on temporal transductive, the recent work ~\cite{siam2022temporal} applies reasoning mechanisms and has achieved good results in cross-domain scenarios. In a word, all the above-mentioned few-shot segmentation methods are only for a single modality - image or video, and cannot handle segmentation under multimodal conditions (i.e., with linguistic referring expressions).\par


\myparagraph{Referring Video Object Segmentation.} 
Gavrilyuk et al.~\cite{gavrilyuk2018actor} first introduce the RVOS task. They generate convolutional dynamic filters from textual representations and convolve them with visual features of different resolutions to obtain segmentation masks.
To overcome the limitations of traditional dynamic convolution, Wang et al.~\cite{wang2020context} propose a context-modulated dynamic convolution operation for RVOS, where the kernel is generated from language sentences and surrounding contextual features. However, since the focus is only on video actors and actions, their approach only applies to a few object classes and action-oriented descriptions.
Weak-Shot Semantic Segmentation (WSSS) ~\cite{chenweak,zhou2021weak} focuses on the overall scene in the image, treating masks and text as the support set and text as the query set. However, in WSSS, the text is limited to single words or phrases indicating class names and directly mapped to labels for pixel-level classification.\par

Khoreva et al.~\cite{khoreva2018rvos} propose a two-stage approach that first performs referring expressions grounding and then utilizes the predicted bounding boxes to guide pixel-wise segmentation. Seo et al.~\cite{seo2020urvos} also present a framework called URVOS, which first predicts the initial mask based on the image and then utilizes the predicted mask of the previous frame for RVOS by memorizing the attention module.\par

Most recent RVOS works employ cross-attention to interact visual images with linguistic information.  LBDT~\cite{ding2022language} uses language as an intermediate bridge to connect temporal and spatial information and leverages cross-modal attention operations to aggregate language-related motion and appearance. 
MMVT~\cite{zhao2022modeling} calculates the optical flow between frames and fuses it as motion information with text features and visual features. 
However, these frame-based spatial granularity multimodal fusion methods have limitations and tend to lead to mismatches between visual and linguistic information. Therefore, a recent piece of work~\cite{wu2022multi,zhao2022modeling} explores a novel multi-level representation learning method and introduces dynamic semantic alignment to adaptively fuse the two modal information.\par

Transformer~\cite{vaswani2017transformer} has been widely applied and achieved great success in many computer vision tasks, such as object detection~\cite{carion2020detr,zhu2020deformable} and image segmentation~\cite{SETR,cheng2021maskformer}. Since DETR~\cite{carion2020detr} introduces a new query-based paradigm, the latest works ~\cite{botach2021mttr, wu2022language, li2022r} prefer to apply the DETR's framework for RVOS task. 
Specifically, they utilize Transformer structures to interact visual images with linguistic data. Because Transformer has a remarkable ability to mine non-local correlation, they are able to attain SOTA performance in accuracy and efficiency.
Despite the relative effectiveness of current RVOS techniques, they are primarily restricted to regular supervised learning settings, which would not be able to deal with unseen scenes with few shots.

\section{Methods}
\label{sec:method}
\subsection{Overview} \label{sec:method overview}
In the setting of FS-RVOS, we have training and testing datasets $D_{train}$ and $D_{test}$ with disjoint category sets $C_{train}$
and $C_{test}$, i.e., $C_{train} \cap C_{test}$ = $\emptyset$. 
Similar to few-shot learning tasks~\cite{snell2017prototypical}, episode training is adopted in this work, where $D_{train}$ and $D_{test}$ consist of several episodes.
Each episode contains a support set $S$ and a query set $Q$, where the text-referred objects (target objects) from both sets belong to the same class.
The support set has $K$ image-mask pairs $S=\left\{x_{k},m_{k} \right\}^{K}_{k=1}$ and the corresponding referring expression with $L$ words $T_s=\left\{ t_i \right\}^{L}_{i=1}$, where $m_k$ is the ground-truth mask of the video frame $x_k$.
The query set $Q=\left\{ x_{i}^{q} \right\}_{i=1}^{N}$ is a selection of consecutive frames from a video and the corresponding natural language description with $M$ words $ T_q=\left\{ t_i \right\}^{M}_{i=1}$, $N$ is the number of frames. With the setting above, FS-RVOS encourages models to segment objects with the unseen class in a query set based on a few samples in the support set. \par 

\begin{figure}[t]
  \centering
  \includegraphics[width=0.99\linewidth]{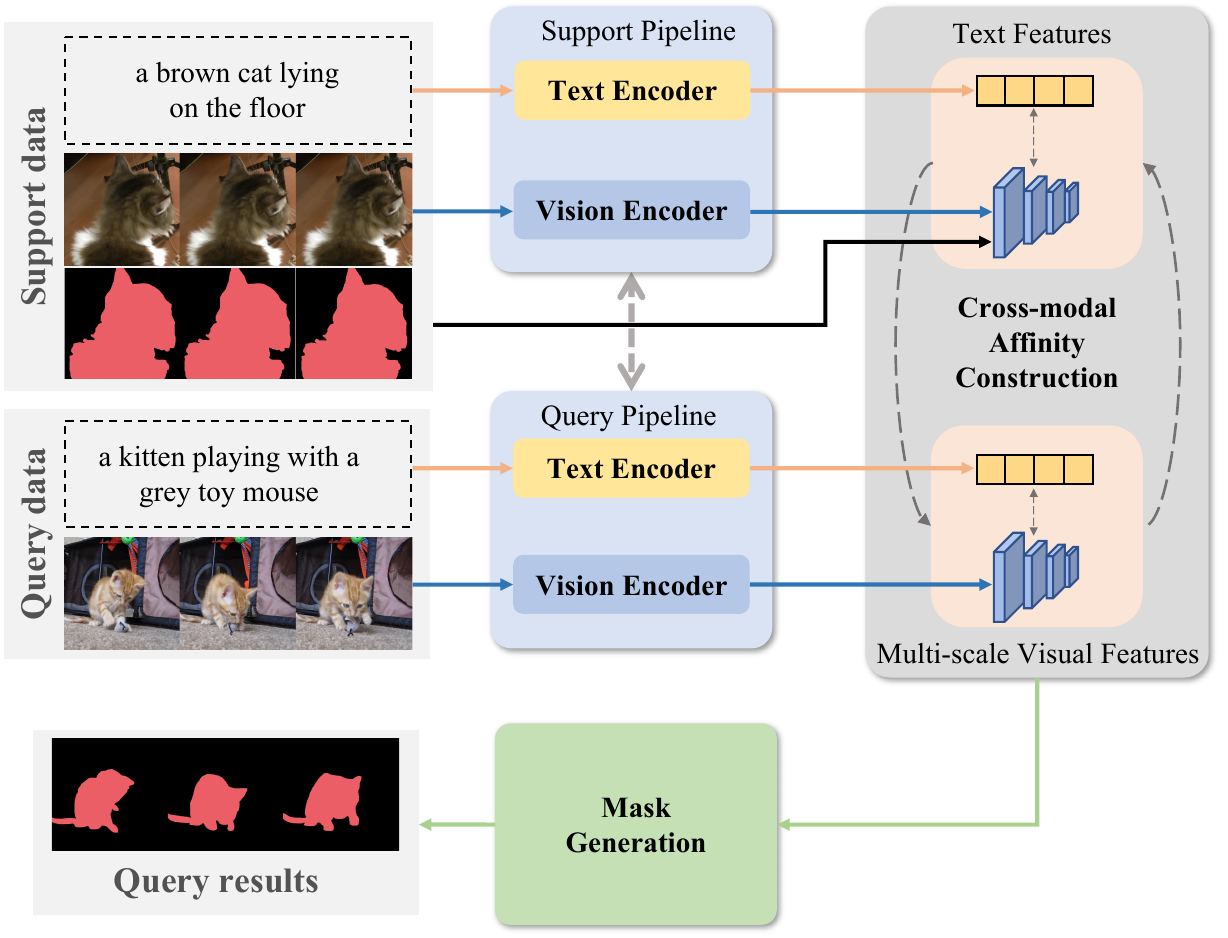}
  \vspace{-1mm}
   \caption{
The overall pipeline of our framework. 
The feature encoder extracts visual and textual information from the support and query sets. Then, the Cross-modal Affinity module calculates the multimodal information affinity between the support set and the query set. Finally, the final segmentation result is obtained through Mask Generation. }
   \label{fig:figure2}
   \vspace{-2mm}
\end{figure}

As shown in Figure~\ref{fig:figure2}, our framework mainly consists of a Feature Extraction module, a Cross-modal Affinity module (CMA), and a Mask Generation module. With the support and query data as input, the framework predicts object masks for the query data, under the guidance of the corresponding language expressions. Specifically, 
the vision and text encoders extract features for visual and textual inputs, respectively. Then, CMA fuses visual and textual features hierarchically. The fused features are used to build the relationship between the support set and the query set. The output features are finally fed to the Mask Generation module to get the final segmentation results.
\par

\subsection{Feature Extraction} \label{sec: feature extraction}
We use a shared-weight visual encoder to extract multi-scale features from each frame in the support set and query set, resulting in visual feature sequences $F_{vs}=\left\{ f_{vs} \right\}_{vs=1} ^K$ and $F_{vq}=\left\{ f_{vq} \right\}_{vq=1}^N$.
For linguistic information, we use a Transformer-based text encoder ~\cite{liu2019roberta} to extract text features $F_{ts}=\left\{ f_{ts} \right\}_{ts=1}^{L}$ and $F_{tq}=\left\{ f_{tq} \right\}_{tq=1}^{M}$ from the natural language descriptions $T_s$ and $T_q$ that correspond to the input support data and the query data.

\subsection{Cross-modal Affinity Construction} \label{sec:cross-modal affinity}

With a few samples in the support set, the goal of FS-RVOS is to efficiently leverage the given information and quickly adapt to relevant scenarios. 
Compared with conventional few-shot tasks, FS-RVOS not only builds the affinity between the support and query sets but also involves multimodal relationships between videos and referring expressions. Therefore, FS-RVOS is more challenging and requires specialized solutions for high-quality results. 

\begin{figure}[t]
  \centering
  \includegraphics[width=\linewidth]{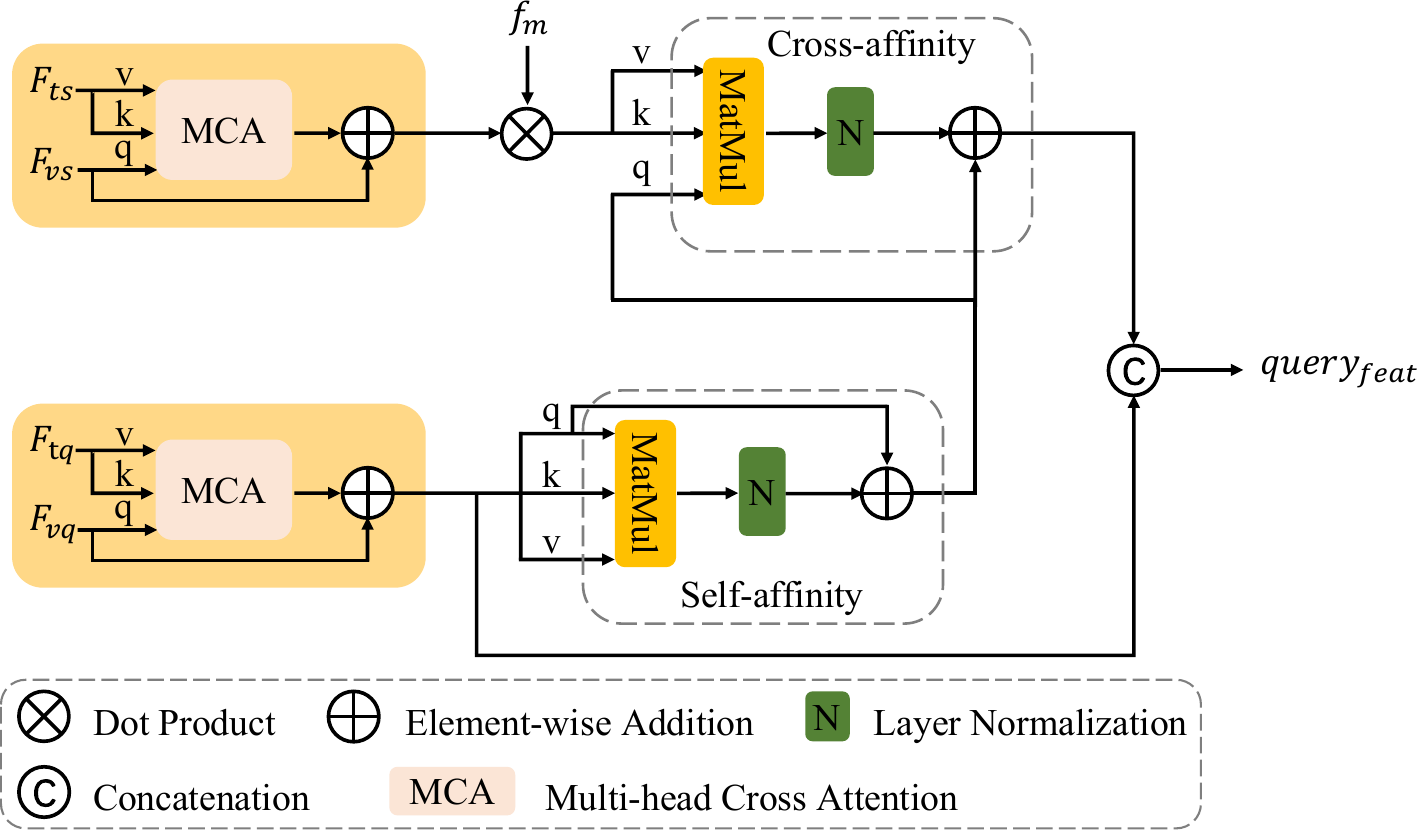}
  \vspace{-4mm}
   \caption{The architecture of the Cross-modal Affinity (CMA) module. We use multi-head cross-attention to fuse visual and text features to get more robust features. Self-affinity for modeling contextual information on query features and cross-affinity for aggregating beneficial information from support features.}
   \label{fig:figure3}
   \vspace{-2mm}
\end{figure}

We propose the Cross-Modal Affinity (CMA) module to achieve this, as shown in Figure~\ref{fig:figure3}. 
We first perform cross-attention fusion between the visual and textual features of support and query data, to obtain pixel-augmented multimodal features.
Then, in order to aggregate beneficial information in support features, we build an affinity relationship between the support set and the query set.\par



Due to the diversity of referred objects and the drastic changes between video frames, it is a challenge to achieve the accurate location of the target only by visual information. 
Therefore, to obtain the accurate positioning of the segmentation target, we use language information as a supplement, which contains a specific description of the referred objects.
To interact and align the visual features and text features, multi-head cross-attention (MCA) is proposed to fuse multimodal information, achieving two multi-scale feature maps $F_{vs}^{'}=\{ f_{s}^{'} \}_{s=1}^K$ and $F_{vq}^{'}=\{ f_{q}^{'} \}_{q=1}^N$:
\begin{equation}
f_{vs}^{'} = \text{MCA}(f_{vs},f_{ts}),\  f_{vq}^{'} = \text{MCA}(f_{vq},f_{tq}),
\end{equation}

\par
\noindent where $f_{vs}$, $f_{vq}$ represent the visual features of support and query, $f_{ts}$, $f_{tq}$ are their corresponding textual features.
Here we calculate an affinity between textual features and visual features to filter out irrelevant visual information. 
Compared with concatenation, MCA suits our framework better since it can leverage the similarities between multimodal features for information complementation. 
\par

The affinity between the support set and query set indicates the multimodal feature correlation among them, providing valuable clues for the segmentation of the query data. 
Although the objects to segment in the support and query sets belong to the same category, their visual properties usually have significant differences, such as appearance, pose, and scene. 
This means only a tiny part of the information in the support data is conducive to segmenting the query data, while other information will cause bad results. 
Therefore, we urgently need to solve the problem of computing the correct affinity relationship between multimodal information in the support and query sets.

To achieve this, we propose a self-affinity block to encode the query features and a cross-affinity block to enable the query features to focus on beneficial pixels in the support features. In particular, given the input query features, we utilize a convolution operation to map them as query $q_q$, key $k_q$, and value $v_q$. We perform the same operation for the support features to map them to key $k_s$ and value $v_s$. 
The input of the self-affinity block does not include the support features and mainly aggregates the context information of the query features for better segmentation.

First, we calculate the affinity map $A^Q = \frac{q_q \cdot (k_{q})^{\text{T}}}{\sqrt{d_{head}}}$, 
where $d_{head}$ is the hidden dimension of the input sequences, and we assume all sequences have the same dimension 256 by default. 
Therefore, the query features after the self-affinity block are represented as:
\begin{equation}
q_s = \text{Softmax}(A^Q)v_{q}.
\end{equation} \par
We then feed the obtained query features to the cross-affinity block. The purpose of the cross-affinity block is to construct the cross-affinity relationship between the support features and the query features and aggregate the useful information. Our cross-affinity block can be formulated as:
\begin{equation}
query_{feat} = \text{Softmax}(\frac{q_s \cdot (k_{s})^{\text{T}}}{\sqrt{d_{head}}})v_{s},
\end{equation}\par
\noindent where $q_s$ is the output of the self-affinity block.
Through these two modules, query features enhance features by modeling contextual information and computing the correlation between support and query features, effectively avoiding the attention bias caused by irrelevant features.
\par

\subsection{Mask Generation} \label{sec: mask}
The goal of the Mask Generation module is to find the most relevant objects and decode the features step by step. To achieve this goal, the structure of Deformable-DETR ~\cite{zhu2020deformable} and feature pyramid ~\cite{lin2017feature} has been used in our work.
We add the corresponding positional encoding to the feature sequence that aggregates beneficial information, which is then sent to the Transformer encoder.
In the decoder part of Transformer, we introduce $N$ learnable anchor boxes as queries to represent the instances of each frame. These queries are replicated as decoder input for all frames and finally converted into instance embeddings by the decoder, resulting in $N_q = T \times N$ predictions.\par

In the feature pyramid network, in order to gradually fuse multimodal features from different layers, the output of the Transformer encoder and the features from the vision encoder are stacked to form hierarchical features. 
 We use $f_v^l$ to represent visual features at each level. First, we downsample the multi-scale visual features, and the time dimension remains unchanged.
Then interact with visual and linguistic features in a cross-attention manner to enhance object pixel features, thereby facilitating mask prediction. The fused features are upsampled to restore the previous shape: 
\begin{equation}
Cross(f_v^l, f_{tq}) = \text{Softmax}(\frac{f_v^l \cdot (f_{tq})^{\text{T}}}{\sqrt{d_{head}}})f_{tq},
\end{equation}\par
\noindent where $f_{tq}$ represents the text features corresponding to the query set. Finally, we pass the features of the last layer through a $3\times3$ convolutional layer to get the final feature maps $F_{seg} = \left\{f_{seg}^t \right\}_{t=1} ^T$, where $f_{seg}^t \in \mathbb{R}^{\frac{H}{4}\times \frac{W}{4} \times C }$.

Furthermore, we construct a set of auxiliary heads to obtain the final object mask across frames. The class head is used to calculate the confidence score $s_t$. The score indicates whether the instance corresponds to the referred object and whether the object is visible in the current frame.
The kernel head is implemented by three consecutive linear layers, which generate the parameters $W=\left\{ w_t\right\}_{t=1}^{N_q}$ of the $N_q$ dynamic kernels. We use them as convolution filters on the feature maps, generating a series of segmentation masks.

\begin{figure*}[ht]
\begin{center}
\includegraphics[width=0.99\textwidth]{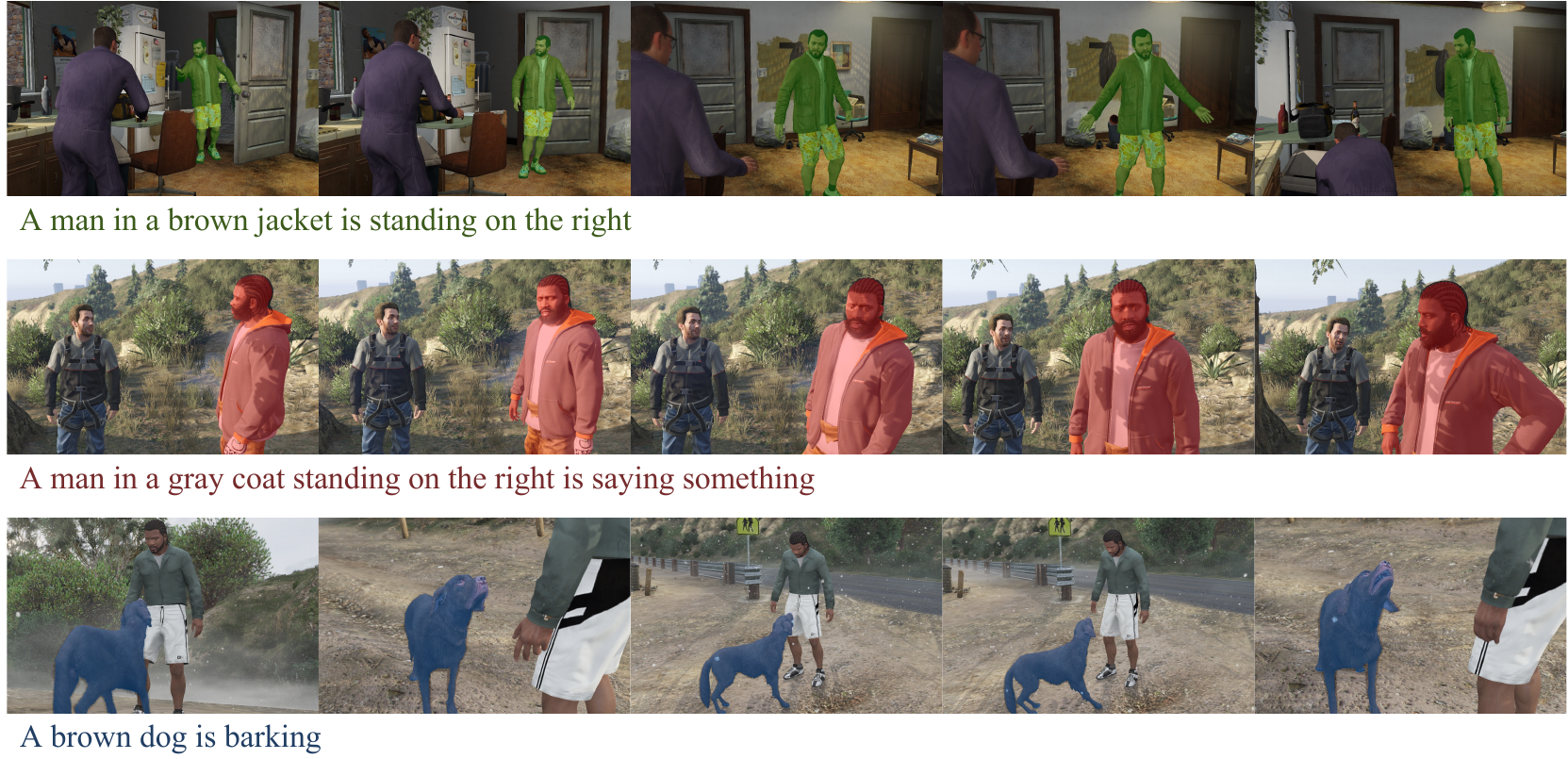}
\end{center}
\vspace{-4mm}
\caption{Annotation examples of the Mini-Ref-SAIL-VOS dataset.}
\label{fig:figure4}
\vspace{-4mm}
\end{figure*}

\subsection{Loss Function} \label{sec: loss}
In Mask Generation, we use $N$ learnable anchor boxes as queries and generate a set of $N_q = T \times N$ prediction sequences.
We denote the predicted sequence as $\hat{y} = \left\{\hat{y}_{i} \right\}_{i=1}^{N_q} $ and the prediction for the $i^{\mathrm{th}}$ instance is expressed as:
\begin{equation}
 \hat{y}_{i}=  \left\{\hat{s}_i^t, \hat{m}_i^t \right\}_{t=1}^T,
\end{equation}

\noindent where $\hat{s}_i^t \in \mathbb{R}^1$ is a fractional score indicating whether the instance corresponds to the referred object. $\hat{m}_i^t \in \mathbb{R}^{\frac{H}{4}\times \frac{W}{4}}$ is the predicted binary segmentation mask. 
Similar to the previous methods~\cite{botach2021mttr,wu2022language}, we use dynamic convolution to generate the object mask. We get the final feature maps $f_{seg}^t$ through the feature pyramid network, and the mask prediction can be calculated by $\hat{m}^t = \left\{ w_{t} *  f_{seg}^t \right\}$.\par
The sequence of ground-truth instances is denoted as $y= \left\{s^t, m^t \right\}_{t=1}^T$, and $s^t$ is an one-hot value and equals 1 when the ground-truth instance is visible in the frame, and 0 otherwise. Our loss function is defined as:
\begin{equation}
\begin{aligned}
    \mathcal{L}(y,\hat{y}_{i})  &=     
    \lambda_{cls}  \mathcal{L}_{cls}(y,\hat{y}_{i}) 
    + \lambda_{kernel} \mathcal{L}_{kernel}(y,\hat{y}_{i}),
\end{aligned}
\end{equation}\par
\noindent where $\lambda_{cls}$, $\lambda_{kernel}$ are hyperparameters to balance the loss. We use focal loss~\cite{lin2017retinanet} (denoted as $\mathcal{L}_{cls}$) to supervise the prediction of instance sequence reference results. $\mathcal{L}_{kernel}$ is a combination of DICE loss~\cite{milletari2016dice} and the binary mask focal loss.\par

\subsection{Inference} \label{sec: inference}
In the inference phase, we treat all input video frames as a whole and predict the mask trajectory for the entire video using only one forward pass. Given an input video and the corresponding linguistic expression, our model generates a sequence of $N$ instances. For each frame, we select the instance sequence with the highest confidence score as the final prediction, and its index can be expressed as:
\begin{equation}
    \overline{s} = \mathop{\arg\max}_{i \in \left \{ 1, 2, ...,N \right \}}s_{i}.
\end{equation}

The final prediction $m=\left\{s_t \right\}_{t=1}^T$ for each frame is obtained from the mask candidate set $\hat{m}^t$ with index $\overline{s}$.



\section{Benchmark}
Since the existing RVOS datasets ~\cite{seo2020urvos,khoreva2018rvos,xu2015a2d,gavrilyuk2018actor} only target specific scenarios, the model cannot handle the diverse scenarios in the real world. Likewise, these datasets are not suitable for the few-shot RVOS problem. This is because the train/test/validation sets of these datasets have class repetition and cannot be used to evaluate the generality of unseen classes.

\myparagraph{Mini-Ref-YouTube-VOS.}
To match the FS-RVOS setting, we build up a new dataset called Mini-Ref-YouTube-VOS based on the Ref-YouTube-VOS dataset~\cite{seo2020urvos}. 
The data that can be directly obtained from the Ref-YouTube-VOS dataset contains 3,471 videos, 12,913 referring expressions, and annotated instances covering more than 60 categories.
However, some videos in this dataset consist of multiple category instances. When preparing data for the few-shot setting, we cleaned up the dataset, i.e., removing such videos and keeping only those containing only one category instance, a total of 2387 videos were obtained.\par

The video data in the dataset should be class-balanced, and the number of samples in each class should not vary too much to avoid overfitting any class. 
Therefore, we deal with the categories whose number of videos does not meet the requirements to ensure the class balance of the dataset.
After the above screening, 1,668 videos were obtained, including 48 classes. To better show the model results, we adopt the cross-validation method to divide the dataset into four folds on average. Each fold contains 36 training and 12 test classes with disjoint categories.

\myparagraph{Mini-Ref-SAIL-VOS.}
Most Mini-Ref-YouTube-VOS data involve natural scenes of a relatively homogeneous type and therefore do not represent the diversity of data in the real world.
To better demonstrate the generalization of our model, we collect videos from SAIL-VOS~\cite{hu2019sail} to construct a new dataset Mini-Ref-SAIL-VOS. The SAIL-VOS dataset is a synthetic dataset collected from a video game GTA-V, aiming to foster semantic amodal segmentation research. 
In SAIL-VOS, each frame is accompanied by densely annotated, pixel-wise
and amodal segmentation masks with semantic labels.
Since the data is collected from the game, phenomena such as shot transition, segmentation target cross-frame, and target occlusion are inevitable, which brings challenges to the segmentation task.\par
We reorganize the SAIL-VOS dataset to pick out samples suitable for the FS-RVOS setting.
First, for videos with few frames for segmentation targets, we directly discard them. For the case where the segmentation target appears across frames, we manually delete the frames where the target does not appear in the middle to maintain the temporal continuity of the segmentation target.
For phenomena such as object occlusion, we characterize it as a challenge and do not deliberately delete video frames with object occlusion. Following the above settings, we collected a dataset with 68 videos and 3 semantic categories.\par
It is worth noting that although there are accurate mask annotations in the SAIL-VOS dataset, natural language description corresponding to the segmentation target is not available. 
Thus, to adapt it to FS-RVOS, we employed expert annotators to provide referring expressions after data collection.
Given a pair of videos for each annotator, the video frames are superimposed with corresponding masks to indicate the objects to be segmented.
The annotators were then asked to provide a distinguishing statement with a word limit of 20 words. To ensure the quality of natural language annotations, all annotations are verified and cleaned up after the initial annotation. The target will not be used if the natural language description cannot clearly describe the target. As shown in Figure~\ref{fig:figure4}, we show the selected videos along with referring expressions.\par


\section{Experiments}

\subsection{Implementation Details} \label{sec: implementation}
We adopt ResNet-50~\cite{he2016resnet} and RoBERTa-Base~\cite{liu2019roberta} as our vision and text encoders, respectively. During the training stage, the parameters of both encoders are frozen. In our experiments, we adopt a 5-shot setting. Specifically, we extract 5 consecutive frames and the corresponding referring expressions from a certain video of a class as a support set. The query set is composed of consecutive frames and corresponding referring expressions extracted from other videos belonging to the same class. We test each fold 5 times and report the average confidence of the results. Our model utilizes AdamW for optimization. The weight decay is $5 \times 10^{-4}$ and the initial learning rate is $1\times 10^{-4}$. To balance GPU memory efficiency, we downsample all video frames, with the shortest video frame size being 360 and the longest 640. The parameters of the loss function are set as $\lambda_{cls}=2$, $\lambda_{kernel}=5$.
All methods for conducting experiments will be pre-processed and fine-tuned in the same way, i.e., pre-trained on Ref-COCO~\cite{yu2016refcoco} dataset.\par
Following the settings of previous RVOS works, we use the region similarity ($\mathcal{J}$) and the contour accuracy ($\mathcal{F}$) to measure the model performance. 

\begin{table}[t]
    \begin{center}
        \setlength{\tabcolsep}{1.3mm}{\begin{tabular}{c |  c | c | c | c | c | c }

\toprule

\multicolumn{2}{c|}{\multirow{1}{*}{Method}}   & \multirow{1}{*}{Fold-1} 
& \multirow{1}{*}{Fold-2} & \multirow{1}{*}{Fold-3} & \multirow{1}{*}{Fold-4}
& \multirow{1}{*}{Mean}  \\

\arrayrulecolor{white}\hline
\arrayrulecolor{black}\hline
\arrayrulecolor{white}\hline
\multirow{2}{*}{$\mathcal{J}$} & DANet~\cite{chen2021delving} & 47 & 33.5 & 38.5 & 44 & 40.8  \\
\multirow{2}{*}{} & Ours & \textbf{59.5} & \textbf{45.3} & \textbf{50.4} & \textbf{57.3} & \textbf{53.1}  \\

\arrayrulecolor{white}\hline
\arrayrulecolor{black}\hline
\arrayrulecolor{white}\hline

\multirow{2}{*}{$\mathcal{F}$} & DANet~\cite{chen2021delving}  & 49.3 & 38.2 & 41.4 & 45.8 & 43.7  \\
\multirow{2}{*}{} & Ours  & \textbf{60.8} & \textbf{48.9} & \textbf{51.3} & \textbf{58.1} & \textbf{54.8}  \\

\arrayrulecolor{white}\hline
\arrayrulecolor{black}\hline
\arrayrulecolor{white}\hline

\end{tabular}}









    \end{center}
    \vspace{-4mm}
    \caption{Quantitative results on Mini-Ref-YouTube-VOS.
    We added visual language fusion modules to DANet.
    }
    \label{tab:table1}
    \vspace{-2mm}
\end{table}

\begin{table}[t]
    \begin{center}
        \setlength{\tabcolsep}{1.3mm}{\begin{tabular}{c |  c | c | c | c | c | c }

\toprule

\multicolumn{2}{c|}{\multirow{1}{*}{Method}}   & \multirow{1}{*}{Fold-1} 
& \multirow{1}{*}{Fold-2} & \multirow{1}{*}{Fold-3} & \multirow{1}{*}{Fold-4}
& \multirow{1}{*}{Mean}  \\

\arrayrulecolor{white}\hline
\arrayrulecolor{black}\hline
\arrayrulecolor{white}\hline
\multirow{2}{*}{$\mathcal{J}$} & DANet~\cite{chen2021delving}  & 54.3 & 47.6 & 30.9 & 30.6 & 40.9  \\
\multirow{2}{*}{} & Ours & \textbf{80.9} & \textbf{80.8} & \textbf{80.1} & \textbf{68.9} & \textbf{77.7}  \\

\arrayrulecolor{white}\hline
\arrayrulecolor{black}\hline
\arrayrulecolor{white}\hline

\multirow{2}{*}{$\mathcal{F}$} & DANet~\cite{chen2021delving}  & 54.1 & 48.4 & 35.2 & 36 & 43.4  \\
\multirow{2}{*}{} & Ours  & \textbf{77.1} & \textbf{77} & \textbf{77.3} & \textbf{67.8} & \textbf{74.8}  \\

\arrayrulecolor{white}\hline
\arrayrulecolor{black}\hline
\arrayrulecolor{white}\hline

\end{tabular}}









    \end{center}
    \vspace{-4mm}
    \caption{Quantitative results on Mini-Ref-SAIL-VOS.
    }
    \label{tab:table2}
    \vspace{-2mm}
\end{table}

\begin{table}[t]
    \begin{center}
        \setlength{\tabcolsep}{1.3mm}{\begin{tabular}{c | c | c |c}

\toprule

\multirow{1}{*}{Method} & \multirow{1}{*}{$\mathcal{J}$} & \multirow{1}{*}{$\mathcal{F}$} & \multirow{1}{*}{$\mathcal{J} \& \mathcal{F}$}  \\

\arrayrulecolor{white}\hline
\arrayrulecolor{black}\hline
\arrayrulecolor{white}\hline

\multirow{1}{*}{LBDT~\cite{ding2022language}} & 27.5 / \underline{42.4} & 36.2 / \underline{37.3} &  31.6 / \underline{39.6}\\

\arrayrulecolor{white}\hline
\arrayrulecolor{black}\hline
\arrayrulecolor{white}\hline

\multirow{1}{*}{ReferFormer~\cite{wu2022language}} & 65.1 / \underline{74.1} & 62.8 / \underline{64.9} & 64.0 / \underline{69.5}\\

\arrayrulecolor{white}\hline
\arrayrulecolor{black}\hline
\arrayrulecolor{white}\hline

\multirow{1}{*}{MTTR~\cite{botach2021mttr}} & 66.5 / \underline{69.7} & 64.9 / \underline{68.1} & 65.7 / \underline{68.9}\\

\arrayrulecolor{white}\hline
\arrayrulecolor{black}\hline
\arrayrulecolor{white}\hline



\multirow{1}{*}{Ours} & \textbf{80.9} & \textbf{77.1} & \textbf{79} \\

\arrayrulecolor{white}\hline
\arrayrulecolor{black}\hline
\arrayrulecolor{white}\hline



\end{tabular}}
    \end{center}
    \vspace{-4mm}
    \caption{Comparison with state-of-the-art methods from RVOS on the Mini-Ref-SAIL-VOS dataset to measure the model's generalization. Underlined scores are
achieved after fine-tuning.}
    \label{tab:table3}
    \vspace{-2mm}
\end{table}

\begin{table}[t]
    \begin{center}
        \begin{tabular}{c | c | c}

\toprule

\multirow{1}{*}{Self-affinity} & \multirow{1}{*}{Cross-affinity} & \multirow{1}{*}{$\mathcal{J} \& \mathcal{F}$}  \\

\arrayrulecolor{white}\hline
\arrayrulecolor{black}\hline
\arrayrulecolor{white}\hline
-\ & -\ & 57.9\\

\arrayrulecolor{white}\hline
\arrayrulecolor{black}\hline
\arrayrulecolor{white}\hline

\multirow{1}{*}{\checkmark} & -\ & 59.2\\

\arrayrulecolor{white}\hline
\arrayrulecolor{black}\hline
\arrayrulecolor{white}\hline

\multirow{1}{*}{\checkmark} & \multirow{1}{*}{\checkmark} & 60.2\\

\arrayrulecolor{white}\hline
\arrayrulecolor{black}\hline
\arrayrulecolor{white}\hline



\end{tabular}
    \end{center}
    \vspace{-4mm}
    \caption{Ablation studies that validate the effectiveness of each component in our CMA. The first result is obtained with our baseline.}
    \label{tab:table4}
    \vspace{-3mm}
\end{table}

\begin{figure*}[ht]
\begin{center}
\includegraphics[width=0.99\textwidth]{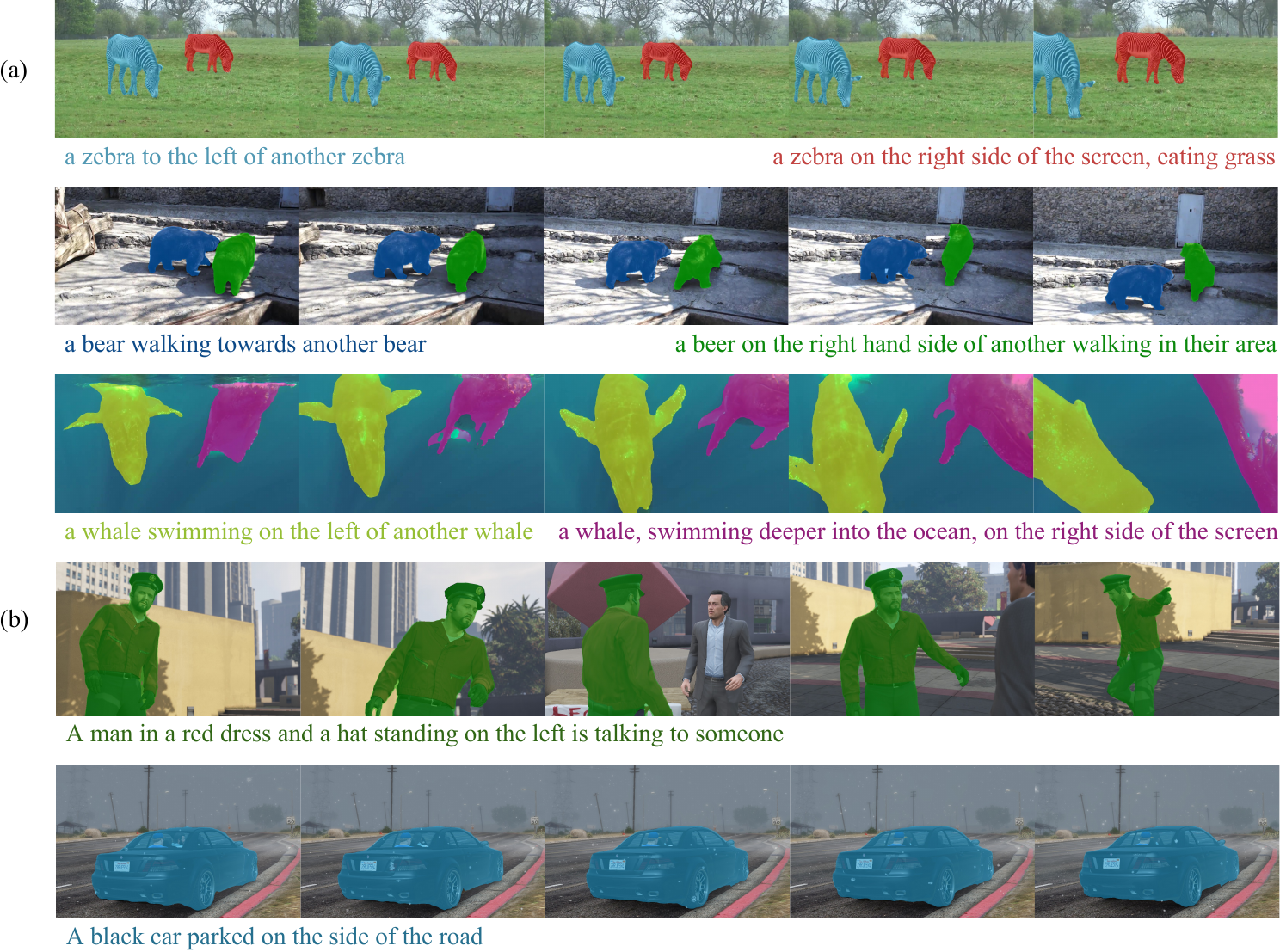}
\end{center}
\vspace{-4mm}
\caption{Qualitative results on (a) Mini-Ref-YouTube-VOS and (b) Mini-Ref-SAIL-VOS.}
\label{fig:figure5}
\vspace{-4mm}
\end{figure*}

\subsection{Results} \label{sec: results}
As a novel problem, no relevant works can be directly used for comparisons. Therefore, we choose DANet~\cite{chen2021delving} as the baseline given we both focus on few-shot video segmentation. For a fair comparison, we add a visual-language fusion module to the Few-Shot VOS model.\par


\myparagraph{Mini-Ref-YouTube-VOS.} 
We present the experimental results of our model on the Mini-Ref-YouTube-VOS dataset in Table \ref{tab:table1}. It can be observed that our method significantly outperforms DANet. 
Compared with previous methods, our method achieves a substantial increase in average performance, with an average improvement of more than 10\%. The excellent performance demonstrates the superiority and robustness of our proposed method.
\par

\myparagraph{Mini-Ref-SAIL-VOS.} 
To evaluate the generalization of our model, we make further experiments and comparisons on the Mini-Ref-SAIL-VOS dataset. We do not perform new training on the Mini-Ref-SAIL-VOS dataset but directly test with the model trained on Mini-Ref-YouTube-VOS. Note that the videos in the Mini-Ref-SAIL-VOS dataset are from Game scenes, and they hold a noticeable domain different from the data in the Mini-Ref-YouTube-VOS dataset. In addition, the objects in some videos are occluded and these phenomena make this dataset somewhat challenging. We show the experimental results in Table \ref{tab:table2}. According to the Table, it is clear that our method achieves a significant improvement over the baseline.\par

Moreover, we also make further comparisons with some state-of-the-art RVOS methods ~\cite{ding2022language, wu2022language, botach2021mttr}. Here, we only show the results under Fold-1. To measure the RVOS model's generalization, we first directly test the trained RVOS models on the Mini-Ref-SAIL-VOS dataset.
Since our few-shot learning task needs a few samples as support data, for a fair comparison, these RVOS models will be fine-tuned with a few samples. The corresponding results are shown in Table \ref{tab:table3}. From the table, it can be seen that although the fine-tuning of the existing models can lead to performance gains, there is still a big gap compared to our proposed approach.
This is because our model effectively constructs the multimodal relationship between the support set and the query set so that it can quickly adapt to new scenarios by only using a few samples.
\par


\subsection{Ablation Study} \label{sec: ablation}
In this section, we perform an ablation study on the Mini-Ref-YouTube-VOS dataset to evaluate the design and the robustness of the model. Unless otherwise stated, we only show results under Fold-1.
 Denoting $\mathcal{J} \& \mathcal{F}$ as the average of $\mathcal{J}$ and $\mathcal{F}$, we can use the indicator to show the performance of the model.\par

\myparagraph{Cross-modal Affinty.}
We perform ablation studies on the components of the CMA in Table \ref{tab:table4}. The results of the baseline are shown in the first line. As mentioned before, the baseline refers to directly concatenating the support features and query features into the Mask Generation to obtain the segmentation mask.\par

First, we only utilize the self-affinity block to establish contextual information between pixels to enhance query features. At this time, support features are concatenated with the enhanced query features and then fed to the Mask Generation module. It can be seen that good results are achieved, indicating that the Transformer does work for modeling features and extracting contextual information. By adding the proposed cross-affinity block, the performance can be further improved by 1\%. 
Such comparisons show that the cross-affinity block properly constructs the multimodal relationship between the support set and the query set, effectively avoiding the bias of the query features.


\subsection{Qualitative Results}
The qualitative results of our model are presented in Figure~\ref{fig:figure5}. From the figure, It can be seen that the proposed model can segment the referred objects accurately in a variety of challenging situations. Furthermore, we also show the qualitative results on the Mini-Ref-SAIL-VOS dataset. In general, our model always achieves high-quality results even in the face of samples from different scenes. \par
\vspace{-2mm}
\section{Conclusion}
In this work, we propose CMA to learn multimodal affinity in a few samples to segment diverse data. Further, we generalize it as a few-shot RVOS problem.
We validate our model on the newly constructed datasets - Mini-Ref-YouTube-VOS and Mini-Ref-SAIL-VOS and obtain state-of-the-art performance. We hope this work can cast a light on future FS-RVOS research.

\myparagraph{Acknowledgements.}   This work is supported in part by the National Key R\&D Program of China (Grant No.
2022YFF1202903), the National Natural Science Foundation of China (Grant No. 61971004, 62122035), the Natural Science Foundation of Anhui Province, China (Grant No. 2008085MF190), and the Equipment Advanced Research Sharing Technology Project, China (Grant No. 80912020104).  

{\small
\bibliographystyle{ieee_fullname}
\bibliography{iccv2023_Few_Shot_RVOS/main}
}

\end{document}